\documentclass{article}

\usepackage{xarxiv}

\usepackage{amssymb}
\usepackage{amsthm}
\usepackage{graphicx}
\usepackage{amsmath}
\usepackage{hyperref}
\usepackage{amsfonts,amssymb}
\usepackage{enumerate}

\newtheorem{assumption}{Assumption}
\usepackage{caption}
\usepackage{subcaption}
\usepackage{multirow}
\usepackage[htt]{hyphenat}
\usepackage{color}
\ifodd 1 \newcommand{\rev}[1]{{\color{blue}#1}}
\else \newcommand{\rev}[1]{} \fi

\begin{document}
\title{Conformal Prediction Intervals for Remaining Useful Lifetime Estimation}
\author{Alireza Javanmardi, Eyke H\"ullermeier		
}

\author{
  {\bf Alireza Javanmardi, Eyke H\"ullermeier} \\
  Institute of Informatics\\
  University of Munich (LMU)\\
  \texttt{alireza.javanmardi@ifi.lmu.de, eyke@lmu.de} 
}

\maketitle

\begin{abstract}
The main objective of Prognostics and Health Management is to estimate the Remaining Useful Lifetime (RUL), namely, the time that a system or a piece of equipment is still in working order before starting to function incorrectly. In recent years, numerous machine learning algorithms have been proposed for RUL estimation, mainly focusing on providing more accurate RUL predictions. However, there are many sources of uncertainty in the problem, such as inherent randomness of systems failure, lack of knowledge regarding their future states, and inaccuracy of the underlying predictive models, making it infeasible to predict the RULs precisely. Hence, it is of utmost importance to quantify the uncertainty alongside the RUL predictions. In this work, we investigate the conformal prediction (CP) framework that represents uncertainty by predicting sets of possible values for the target variable (intervals in the case of RUL) instead of making point predictions. Under very mild technical assumptions, CP formally guarantees that the actual value (true RUL) is covered by the predicted set with a degree of certainty that can be prespecified. We study three CP algorithms to conformalize any single-point RUL predictor and turn it into a valid interval predictor. Finally, we conformalize two single-point RUL predictors, deep convolutional neural networks and gradient boosting, and illustrate their performance on the Commercial Modular Aero-Propulsion System Simulation (C-MAPSS) data sets.  
\end{abstract}


\section{Introduction}\label{sec:intro}
Prognostics and Health Management (PHM) is devised to monitor the health state of industrial components and conduct maintenance operations when necessary. It can noticeably increase the efficiency of industrial assets by reducing their downtime, maintenance frequency, and costs accordingly. One essential element of the PHM is the Remaining Useful Lifetime (RUL) estimation, which refers to predicting the amount of time left before a system stops working as required \cite{Jardine2006}. 

In the past years, many data-driven approaches have been proposed for this problem \cite{Lei2018a}. From a machine learning perspective, these works attempt to solve a regression problem, that is, to discover the relationship between the condition monitoring (CM) data and the RUL. The main objective of most of these works is to predict the RUL as accurately as possible. Obviously, due to the noisiness of CM data, the stochastic behavior of systems failure, the unpredictability of systems' future states, and even the imprecision of the regression models, the RUL estimation problem is heavily affected by uncertainty \cite{Sankararaman2015}. Moreover, RUL predictions will eventually affect the maintenance process, which is a delicate decision-making procedure. Hence, as a prerequisite for reliable employment in industry, it is essential to equip such predictions with a valid representation of their confidence, for instance, by answering questions of the following kind: How confident is the model with the prediction it made? What is the probability that the true RUL will actually be shorter than the predicted value, and if so, by what amount? 

Uncertainty quantification (UQ) is a field in machine learning that deals with questions like these \cite{Hullermeier2021}. In the regression problem, one way to quantify uncertainty is to predict an interval equipped with a level of confidence instead of a single value. A prediction interval provides a lower and an upper bound on the target variable, i.e., the RUL in our problem. Ideally, prediction intervals are as short as possible, and their length should represent the difficulty of the prediction (\textit{adaptivity} property); in other words, the harder the prediction for a given data point, the higher the uncertainty and the wider the interval. More importantly, a predicted interval ought to contain the actual value of the response variable with a certain degree of probability (\textit{coverage} property) \cite{Angelopoulos2021}. 

Conformal prediction (CP) is a framework for constructing prediction intervals, or, more generally, prediction regions, which has gained increasing interest in the recent past \cite{Vovk2005a}. 
CP can be applied in a very versatile way and guarantees coverage property under mild technical assumptions. It is non-parametric, i.e., it makes no specific distributional assumptions on the data-generating process. 
Moreover, it can be put on top of any single-point RUL predictor, thereby turning it into an interval predictor. In this work, we employ the conformal prediction framework for the RUL estimation problem. 
More specifically, we present three CP methods and describe how to turn any single-point RUL estimator into a valid interval predictor using any of them. To the best of our knowledge, this paper is the first to investigate the CP framework for the RUL estimation problem. 

The paper is organized as follows. After a brief overview of related work, the RUL estimation problem is defined in Section \ref{sec:problem}. This is followed by a presentation of the conformal prediction framework and its two variants for interval prediction. In Section \ref{sec:exp}, two single-point RUL predictors are conformalized using the CP methods, and their performance is evaluated experimentally using the C-MAPSS data set.

\section{Background and Related Work}\label{sec:related}

\subsection{RUL Estimation}
RUL estimation methods can be categorized into model-based and data-driven approaches \cite{An2014}. Model-based approaches specify a physical degradation model according to prior domain knowledge of the system and utilize historical data to identify its parameters. Data-driven approaches employ data to discover relationships between system state and failure. Corresponding approaches range from classical machine learning methods such as support vector machines (SVM) \cite{Benkedjouh2013}, K-nearest neighbors (KNN) \cite{Mosallam2016}, random forests (RF) \cite{Zhang2017}, and gradient boosting (GB) \cite{Zhang2017} to modern deep learning techniques, including deep belief networks (DBN) \cite{Zhang2017}, deep convolutional neural networks (DCNN) \cite{Babu2016, Li2017}, recurrent neural networks (RNN) \cite{Heimes2008} and its variants \cite{Wu2017, Chen2019, Elsheikh2019}. More recently, the problem has also been tackled by means of automated machine learning (AutoML) \cite{Tornede2020AutoML, Tornede2021coevolution}. 


\subsection{Uncertainty Quantification in Regression}

As pointed out by the authors in \cite{Sankararaman2015}, due to the existence of multiple sources of uncertainty in prognostics, predicting a single value as an RUL is not very meaningful. Nevertheless, uncertainty quantification is relatively understudied in the field of data-driven RUL estimation. From a machine learning point of view, RUL estimation is considered a regression problem, for which four fundamental classes of interval predictors exist in the literature: Bayesian methods, ensemble methods, direct interval prediction, and CP methods \cite{Dewolf2022}. 

The most common Bayesian methods are Gaussian processes \cite{Liu2019, Wu2020, Biggio2021} and Bayesian neural networks \cite{Peng2020, Benker2021}. Their basic idea is to adopt a prior distribution over model parameters, which is then turned into a posterior distribution in the light of observed data. The advantage of these methods is their theoretical soundness and formal guarantees, while their main weakness is their vulnerability to model misspecification.

In ensemble methods, multiple machine learning models are trained simultaneously, and the statistics of their predictions (e.g., mean and variance) are utilized to quantify uncertainty \cite{Rigamonti2018, Liao2018}. Broadly speaking, the more the predictions diverge, the higher the uncertainty seems to be. Ensemble methods are simple and efficient, but somewhat ad-hoc and difficult to interpret (e.g., they do normally not support a probabilistic interpretation).

The most well-known direct interval predictors are quantile regression models that construct intervals by providing lower and upper quantiles of response variables given their features \cite{Zhao2020}. However, the intervals constructed by estimated quantiles do not guarantee coverage when dealing with a finite number of samples \cite{Romano2019}. 

\subsection{Conformal Prediction}

Conformal prediction delivers reliable predictions in the form of sets or prediction regions (intervals in the case of regression), which are guaranteed to contain the sought target value with a predefined level of confidence. CP assumes no specific distribution for the data and works under the mild assumption of data exchangeability, which requires the joint probability distribution of a set of data points to be independent of their order \cite{Angelopoulos2021}. CP has originally been introduced for transductive inference in an online setting, but inductive variants have been developed later on\,---\,we refer to \cite{Lei2018} for details. Due to the computational complexity of full (transductive) CP, we focus on the inductive variant of so-called split conformal prediction in this paper. 

Split CP divides the training data into subsets for proper training and calibration, using the former to fit the regression model and the latter to quantify the uncertainty of predictions (in the test set). One major drawback of the original split CP algorithm is that, in the case of regression, the prediction intervals have the same length for all data points in the test set. 
In \cite{Romano2019} this issue is addressed by fitting two quantile regression models (for lower and upper bounds on the target values, respectively) and calibrating them using the calibration subset to ensure the coverage property. 

All theoretical guarantees of the aforementioned approaches rely on the exchangeability assumption. However, this assumption can easily be violated, especially when dealing with ordered data such as time series. 
In \cite{FoygelBarber2022}, this issue is tackled by assigning weights to calibration data points based on their ``similarity'' to each data point in the test set.
We will discuss this approach in more detail in Section \ref{sec:problem:nex-SCP}.

\section{Conformal Prediction for RUL Estimation}\label{sec:problem}

In PHM, each data instance is represented by a (possibly multivariate) time series $\mathbf{Z}_i := \lbrace z_1^{(i)}, z_2^{(i)}, \ldots, z_{T_i}^{(i)}\rbrace$, a collection of condition monitoring data from the moment system $i$ starts operating up to time $T_i$, and a scaler $F_i$ indicating its failure time. The RUL of instance $i$ at time $t$ can be computed as
\begin{align}\label{eq:RUL}
    y_t^{(i)} = F_i - t, 
\end{align}
$t\in[T_i] := \{1, \ldots, T_i\}$. 
Typically, for training data $ \lbrace (\mathbf{Z}_i, F_i)\rbrace_{i=1}^{N_\text{train}}$, the time series terminate when a failure occurs, i.e., $F_i = T_i$. Such data is also referred to as \textit{run-to-failure} data. On the other hand, for a test data set $\lbrace (\mathbf{Z}_k, F_k)\rbrace_{k=1}^{N_\text{test}}$, the series may also end at a random time before a failure occurs, i.e., $F_k \geq T_k$. Clearly, the ultimate objective is to estimate $y_{T_k}^{(k)}$ for every data point $(Z_k, F_k)$ in the test data set.  

Depending on the regression algorithm, these training and test data sets usually need to be transformed into 
\begin{align*}
    \mathcal{D} = \big\{ (x_i, y_i)\big\}_{i=1}^{N'_\text{train}}
\end{align*}
and
\begin{align*}
    \mathcal{D}_\text{test} = \big\{ (x_k, y_k)\big\}_{k=1}^{N'_\text{test}} \, ,
\end{align*}
correspondingly. In the simplest case, the transformed data set is simply the collection of all CM data from all instances and their corresponding RULs, e.g., 
\begin{align}\label{eq:transformed_data:all}
    \mathcal{D} =\left\{ (z_t^{(i)},y_t^{(i)}) : i\in [N_\text{train}],~t\in[T_i]\right\} \, .
\end{align}
The authors of \cite{Li2017} consider a time window of length $L_w$ and stack all CM data within that interval to construct two-dimensional features. As a result, the transformed data set can be written as 
\begin{align}\label{eq:transformed_data:window}
    \mathcal{D} =\bigg\lbrace \bigg([z_{t - L_W+1}^{(i)}, \ldots, z_{t-1}^{(i)}, z_{t}^{(i)}]^\intercal,y_t^{(i)}\bigg)   
    : i\in [N_\text{train}],~L_w<t\leq T_i\bigg\rbrace \, ,
\end{align}
where $[\cdot]^\intercal$ denotes the transpose of $[\cdot]$.

Regardless of the choice of data transformation method, in the conventional regression problem, the objective is to train a model $M$ on $\mathcal{D}$ that mimics the relationship between instances (independent variables) $x_i$ and targets (dependent variables) $y_i$, so that $\hat{y}_\text{new} := M(x_\text{new})$ is close to $y_\text{new}$ for every new pair $(x_\text{new}, y_\text{new}) \in \mathcal{D}_\text{test}$. Alternatively, conformal prediction attempts to provide an interval $C^{\alpha} (x_\text{new}) \subset \mathbb{R}_{\geq 0}$ that contains $y_\text{new}$ with a user-defined \textit{coverage rate} $1-\alpha$, where $\alpha$ is the \emph{error rate}. Thus, the following \textit{coverage property} holds: 
\begin{align}\label{eq:coverage}
    \mathbb{P}\bigg(y_\text{new} \in C^{\alpha} (x_\text{new})\bigg) \geq 1 - \alpha.
\end{align}

\begin{figure}[t]
\centering
\includegraphics[width = 0.8\columnwidth]{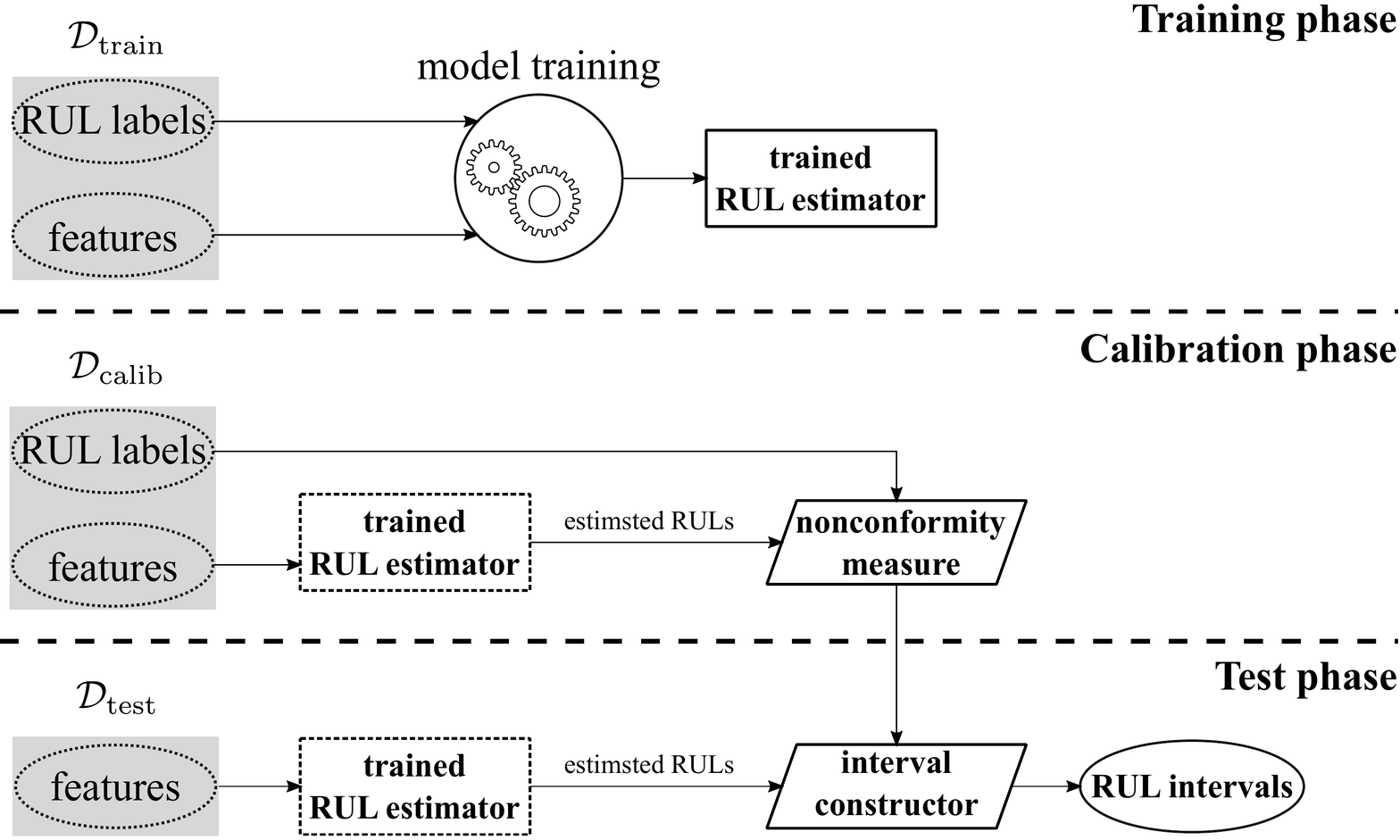}
\caption{The general procedure of split conformal prediction for RUL estimation problem.}
\label{fig:cp}
\end{figure}

In this paper, we concentrate on the split conformal prediction (SCP) framework and two of its variants, namely conformalized quantile regression (CQR) and non-exchangeable split conformal prediction (nex-SCP). The general procedure of these methods is depicted in Fig.\ \ref{fig:cp}.

\subsection{Split Conformal Prediction Framework}\label{sec:problem:SCP}

As its name suggests, SCP starts by randomly splitting the original training data $\mathcal{D}$ into two disjoint subsets: a proper training set $\mathcal{D}_\text{train}$ and a calibration set $\mathcal{D}_\text{calib}$.
A predictive model $M$ is then trained on the proper training set and used to obtain non-conformity scores for the examples in the calibration data. To this end, CP needs a \textit{non-conformity measure} (aka non-conformity score function) $f$ that takes a tuple $(x,y)$ as input and returns a real-valued score $S= f(x,y)$ as output. The latter is meant to indicate the ``strangeness'' of the data point $(x,y)$, i.e., to measure how non-conforming $(x,y)$ is with the model $M$\,---\,in regression, a natural measure of non-conformity is the absolute residual error between the prediction $\hat{y}_j := M(x_j)$ and the observed outcome $y_j$:
    \begin{align}\label{eq:SCP:non-conformity}
        S_j  = f(x_j,y_j) = |y_j - \hat{y}_j| = |y_j - M(x_j)| \, .
    \end{align}
 
Applying the non-conformity measure $f$ to each data point in $\mathcal{D}_\text{calib}$, one obtains a set of non-conformity scores 
$$
\lbrace S_j : (x_j, y_j) \in \mathcal{D}_\text{calib}\rbrace \, .
$$
Let $q$ be the $\lceil (1+|\mathcal{D}_\text{calib}|)(1-\alpha)\rceil$ smallest value of these scores, where $\lceil \cdot \rceil$ is the ceiling function. Equivalently, $q$ is the $(1-\alpha)$-quantile of the empirical distribution of non-conformity scores 
$$
\frac{1}{|\mathcal{D}_\text{calib}|+1} \, \delta_{+\infty} + \sum_{(x_j,y_j)\in\mathcal{D}_\text{calib}}\frac{1}{|\mathcal{D}_\text{calib}|+1} \, \delta_{S_j} \, , 
$$
with $\delta_x$ denoting the point mass at point $x$. This value can be interpreted as follows: With high probability (namely, a probability of at least $1-\alpha$), the non-conformity of a ``real'' data point $(x,y)$, i.e., a data point sampled from the true underlying distribution, is $\leq q$. 

The basic idea of CP, then, is to ``reject'' any hypothetical data point $(x,y)$ the non-conformity of which exceeds $q$. In accordance with the logic of statistical hypothesis testing, the probability of erroneously rejecting a real data point (conducting a mistake of type 1) is upper-bounded by $\alpha$. More specifically, given a query instance $x_{new}$ for which a prediction is sought, a prediction region $C^{\alpha}(x_\text{new})$ is constructed by testing the hypothesis $(x_{new}, y_{new}) = (x_{new}, y)$ for each candidate value $y \in \mathbb{R}$, and only including those candidates for which this hypothesis cannot be rejected. 
In the case of regression with non-conformity measure (\ref{eq:SCP:non-conformity}), this simply leads to the interval
\begin{align}\label{eq:SCP:interval}
        C_\text{SCP}^{\alpha} (x_\text{new}) = \big[ \, \hat{y}_\text{new} - q, \hat{y}_\text{new} + q \, \big] \, . 
    \end{align}
 
The SCP procedure can be summarized as follows: 
\begin{enumerate}
\def\makelabel{Step~}
    \item \label{SCP:alg:train} A regression model $M$ is trained on $\mathcal{D}_\text{train}$, using any regression algorithm $\mathcal{A}$:
    \begin{align*}
        M \leftarrow \mathcal{A}(\mathcal{D}_\text{train})
    \end{align*}
    \item \label{SCP:alg:calib} The non-conformity scores $S_j$ are computed for all $(x_j, y_j) \in \mathcal{D}_\text{calib}$, using the non-conformity measure (\ref{eq:SCP:non-conformity}), which is the most common measure for regression. 
    \item \label{SCP:alg:interval}The critical non-conformity $q$ is obtained as described above, and a prediction interval 
    $$
    C_\text{SCP}^{\alpha} (x_\text{new}) = [\hat{y}_\text{new} \pm q ] = [M(x_\text{new}) \pm q ]
    $$ 
    is constructed according to (\ref{eq:SCP:interval}) for each $x_\text{new}$ in the test data $\mathcal{D}_\text{test}$. 
\end{enumerate}

It can be easily observed that a lower value of $\alpha$ results in a higher value of $q$ and, consequently, wider intervals.

One limitation of this procedure is a lack of adaptivity: The length of prediction intervals is always the same, namely $2q$, regardless of the query instance $x_\text{new}$. In contrast, one would intuitively expect that the widths of an interval is adapted to the ``difficulty'' of the prediction. 
One way to tackle this problem is to modify the non-conformity measure to account for the difficulty of the data points \cite{Lei2018}. For example, a \textit{normalized} non-conformity measure can be defined as
    \begin{align}\label{eq:normalized:non-conformity}
        S_j = \frac{|y_j - \hat{y}_j|}{\sigma(x_j)} 
    \end{align}
    for calibration data $(x_j, y_j) \in \mathcal{D}_\text{calib}$, where $\sigma$ is another regression model trained on $\lbrace (x_i,|y_i - \hat{y}_i|) : (x_i, y_i) \in \mathcal{D}_\text{train}\rbrace$. Given $q$ as the $\lceil (1+|\mathcal{D}_\text{calib}|)(1-\alpha)\rceil$ smallest value of the normalized non-conformity scores, the prediction interval at $x_\text{new}$ is formed as
    \begin{align}\label{eq:normalized:interval}
         \big[ \, \hat{y}_\text{new} \pm  q\sigma(x_\text{new}) \, \big] \, . 
    \end{align}
    In the following section, we introduce another approach that achieves adaptivity in a different way.

\subsection{Conformalized Quantile Regression Framework}\label{sec:problem:CQR}

The general procedure of the CQR approach is similar to SCP except for the choice of the regression algorithm and the non-conformity measure. Conventional regression algorithms estimate the conditional mean of the response variable given its features (i.e., $\mathbb{E}[Y|X=x]$). In $\tau-$quantile regression, on the other hand, we are interested in estimating a conditional quantile of the response variable given its features (i.e., $Q_\tau(Y|X=x)$\footnote{$Q_\tau(Y|X=x) := \inf\lbrace y\in \mathbb{R} :F(y|X=x) \geq \tau \rbrace$ where $F(y|X=x) := \mathbb{P}\big(Y\leq y|X=x\big)$ is the conditional distribution of $Y$ given $X=x$.
}). This can be accomplished by training a regression model on a specific loss function, the \textit{pinball loss}, wich is defined as follows:
\begin{align}\label{eq:pinball}
    PL_\tau (y,\hat{y}) := \begin{cases}
    \tau (y-\hat{y}) & \text{if } y>\hat{y}\\
    (1-\tau) (\hat{y}-y) & \text{otherwise}
    \end{cases} \, .
\end{align}

The procedure of CQR is as follows: 
\begin{enumerate}
\def\makelabel{Step~}
    \item Two quantile regression models are fitted on $\mathcal{D}_\text{train}$ using any quantile regression algorithm $\mathcal{A}$, one with $\tau_\text{low} = \alpha$ and the other one with $\tau_\text{high} = 1- \alpha$:
    \begin{align*}
        \hat{Q}_{\tau_\text{low}}, \hat{Q}_{\tau_\text{high}} \leftarrow \mathcal{A}(\mathcal{D}_\text{train})
    \end{align*}
    \item For measuring the non-conformity, the following scoring function is used: 
    \begin{align}\label{eq:CQR:non-conformity}
        S_j = \max\big\{ \hat{Q}_{\tau_\text{low}} (x_j) - y_j, y_j - \hat{Q}_{\tau_\text{high}}(x_j) \big\} \, .
    \end{align}
    \item Let $q$ be the $\lceil (1+|\mathcal{D}_\text{calib}|)(1-\alpha)\rceil$ smallest value of the non-conformity scores on the calibration data. The prediction interval for a new test instance $x_\text{new}$ is then constructed as follows:  
    \begin{align}\label{eq:CQR:interval}
        C_\text{CQR}^{\alpha} (x_\text{new}) = \big[ \, \hat{Q}_{\tau_\text{low}} (x_\text{new}) - q, \hat{Q}_{\tau_\text{high}} (x_\text{new}) + q \,  \big] \, . 
    \end{align}
\end{enumerate}
The scoring function \eqref{eq:CQR:non-conformity} has meaningful properties: The score is positive if the actual response variable $y_j$ is outside the interval $[\hat{Q}_{\tau_\text{low}}(x_j), \hat{Q}_{\tau_\text{high}}(x_j)]$, accounting undercoverage; otherwise, if $y_j$ is covered by $[\hat{Q}_{\tau_\text{low}}(x_j), \hat{Q}_{\tau_\text{high}}(x_j)]$, the score is non-positive, thereby handling the overcoverage problem \cite{Romano2019}. 
Furthermore, the length of $C_\text{CQR}^{\alpha} (x_\text{new})$ varies with $x_\text{new}$, which represents adaptivity. It is proved in \cite[Theorem~$1$]{Romano2019} that under the exchangeability assumption of the data points in $\mathcal{D}_\text{calib} \cup \mathcal{D}_\text{test}$, the intervals given in \eqref{eq:CQR:interval} satisfy the coverage property \eqref{eq:coverage}.

\subsection{Non-exchangeable Split Conformal Prediction}\label{sec:problem:nex-SCP}

The coverage property of CP, namely,
\begin{align*}
    \mathbb{P}\bigg(y_\text{new} \in C^{\alpha} (x_\text{new})\bigg) \geq 1 - \alpha,
\end{align*}
is guaranteed under relatively mild technical assumptions \cite{Lei2018}. One important assumption that needs to be satisfied, however, is the exchangeability of the underlying data-generating process.



\medskip

\begin{assumption}[Exchangeability]
Random variables $V_1,V_2,\ldots,V_m$ are called exchangeable if their joint distribution does not depend on their order: 
\begin{align*}
    \mathbb{P}(V_1,V_2,\ldots, V_m) = \mathbb{P}(V_{\pi(1)},V_{\pi(2)},\ldots,V_{\pi(m)})
\end{align*}
for any permutation $\pi:[m]\rightarrow [m]$. 
\end{assumption}
In the case of conformal prediction, the random variables of interest include the data $v_j = (x_j, y_j)$ in the calibration data $\mathcal{D}_\text{calib}$ and the new test case $v_{new} = (x_{new}, y_{new})$.



Exchangeability is weaker than the common assumption of independent and identically distributed (i.i.d.) data, i.e., the former implies the latter but not the other way around. Nevertheless, in the case of data with a temporal component, even exchangeability will probably be violated. In the following, we describe a method proposed in \cite{FoygelBarber2022}, which modifies the SCP framework to give valid intervals even when the exchangeability assumption does not hold.

Consider a new data point $(x_\text{new}, y_\text{new})$ in the test set and assume that data points in $\mathcal{D}_\text{calib} \cup \{ (x_\text{new}, y_\text{new}) \}$ are not exchangeable. Suppose we have an idea about the underlying similarity between the distributions of $(x_\text{new}, y_\text{new})$ and the points in $\mathcal{D}_\text{calib}$. In that case, we can assign weights $w_j \in [0,1]$ for every $(x_j,y_j)$ in $\mathcal{D}_\text{calib}$, with higher weights indicating higher similarity. For instance, consider a time series 
$$
(x_1,y_1), (x_2,y_2),\ldots,(x_t,y_t),(x_{t+1},y_{t+1}) \, ,
$$
with the first $t$ points being the calibration data and $(x_{t+1},y_{t+1})$ the test point. It would then be natural to choose weights $w_1\leq w_2\leq\ldots\leq w_t$ such that the more recent data points have greater weights than the less recent ones. 

Given calibration data $(x_j, y_j) \in \mathcal{D}_\text{calib}$ with non-conformity scores $S_j$ according to \eqref{eq:SCP:non-conformity} and weights $w_j$, we define normalized weights  
\begin{align}\label{eq:normalized_weights}
  \Tilde{w}_j = \frac{w_j}{1 + \sum_{j \in \mathcal{D}_\text{calib}} w_j} \, .
\end{align}
These normalized weights can be used to modify the empirical distribution of non-conformity scores as follows:
\begin{align}\label{eq:beyond:non-conformity_distribution}
    \sum_{(x_j, y_j) \in \mathcal{D}_\text{calib}} \Tilde{w}_j \delta_{S_j} + \Tilde{w}_{+\infty} \delta_{+\infty} \, ,
\end{align}
with $\delta_x$ being the point mass at $x$, and 
$$
\Tilde{w}_{+\infty} = \frac{1}{1 + \sum_{j \in \mathcal{D}_\text{calib}} w_j} \, .
$$    
    The procedure of nex-SCP  is similar to SCP except for the last step. This time, $q(x_\text{new})$ needs to be calculated for every $(x_\text{new}, y_\text{new}) \in \mathcal{D}_\text{test}$ separately and is defined as the $(1-\alpha)$-quantile of the empirical distribution of non-conformity scores given in \eqref{eq:beyond:non-conformity_distribution}. Accordingly, the prediction interval at $x_\text{new}$ is constructed as follows:  
    \begin{align}\label{eq:beyond:interval}
        C_\text{nex-SCP}^{\alpha} (x_\text{new}) = \big[ \, \hat{y}_\text{new} - q(x_\text{new}), \hat{y}_\text{new} + q(x_\text{new}) \, \big] \, .
    \end{align}
Without the assumption of exchangeability, the level of coverage that can be guaranteed is reduced compared to the exchangeable case; we refer to \cite[Theorem~2a]{FoygelBarber2022} for more details on the theoretical properties of this method.

\section{Experiments}\label{sec:exp}
The main focus of this paper is on the CP algorithms and how to conformalize any single-point RUL estimator, turning it into a reliable interval predictor. For this purpose, we employ two existing single-point RUL estimators, Deep Convolutional Neural Networks (\textbf{DCNN}) and Gradient Boosting (\textbf{GB}), and conformalize them using \textbf{SCP}, SCP with normalized non-conformity measure (\textbf{SCP+NNM}), \textbf{nex-SCP}, nex-SCP with normalized non-conformity measure (\textbf{nex-SCP+NNM}), and \textbf{CQR}. Our implementation code is publicly available on  \href{https://github.com/alireza-javanmardi/conformal-RUL-intervals.git}{GitHub}\footnote{\url{https://github.com/alireza-javanmardi/conformal-RUL-intervals}} to enable the reproducibility of the presented results.

\subsection{C-MAPSS Data}
C-MAPSS is a software written in the MATLAB-Simulink environment, which is capable of simulating a large commercial turbofan engine with various tunable input parameters to specify numerous operational profiles, environmental conditions, initial wear degrees, degradations, etc. 
The authors of \cite{Saxena2008} ran this simulation environment multiple times with different parameter values while collecting noisy data from many units, including twenty-one sensor measurements and three operational settings. They provided four data sets of multivariate time series where each data set consists of a training and a test set. At the beginning of each time series, the engine operates normally, and at a random point during the series, it starts to degrade. For training sets, a time series terminates when the engine failure occurs, and the RUL at each time step is defined as the number of time steps left until the end of the series. For the test sets, a series ends at a random time before the failure, while the actual RUL for its last time step is provided. 

\subsubsection{Data Preprocessing}
The details of the four data sets of C-MAPSS are provided in Table \ref{tab:c-mapss}. For data sets $\#1$ and $\#3$, the operating condition does not vary with time (stationary case), while it alters between $6$ different operating modes for data sets $\#2$ and $\#4$ (nonstationary case). Henceforth, we use different data normalization methods for each case. We use a min-max scaler for stationary cases to map the sensor measurements into the range $[-1,1]$. For nonstationary cases, on the other hand, we first apply the K-means clustering algorithm to the 3-dimensional operating settings to cluster them into $K=6$ clusters (operating modes). Six separate min-max scalers (with the range $[-1,1]$) are then used to normalize sensor measurements within each operating mode.

Moreover, seven of the twenty-one sensor measurements have zero (or close to zero) variances and provide no useful information. Hence, we remove these seven sensors with the indices $1$, $5$, $6$, $10$, $16$, $18$, and $19$.

\begin{table}[t] \small  
	\begin{center}  
		\caption{Description of the four C-MAPSS data sets.}
	\label{tab:c-mapss}
	\medskip
	\begin{tabular}{ l c c c c }

		\textbf{Data set}	& \textbf{\# 1} & \textbf{\# 2}	& \textbf{\# 3} & \textbf{\# 4} \\ 
		\hline \hline
        \# Training instances	& 100 & 260	& 100 & 249 \\ 
        \# Test instances 	& 100 & 259	& 100 & 248 \\ 
        \# Operating conditions	& 1 & 6	& 1& 6 \\ 
        \# Fault modes	& 1 & 1	& 2& 2 \\\hline
	\end{tabular}

	\end{center}
\end{table}

A piecewise linear definition of RUL labels (aka rectified labels) exists for this data set that limits the maximum value of the RUL \cite{Heimes2008}. These rectified labels can be easily defined by modifying \eqref{eq:RUL} as follows:
\begin{align}\label{eq:rectified_RUL}
    y_t^{(i)} = \max\bigg(\text{RUL}_{\max}, F_i - t \bigg) 
\end{align}
for all $i$ and $t\in[T_i]$, where $\text{RUL}_{\max}$ is a fixed value chosen on the basis of the observations.
The intuition behind this definition is that the system's degradation begins after a certain degree of usage. Similar to \cite{Li2017}, we also set $\text{RUL}_{\max}$ to $125$ for all four data sets. 
\subsection{Learning Algorithms}
The following single-point RUL estimators are used in our experiments: 
\begin{itemize}
    \item Deep Convolutional Neural Networks (\textbf{DCNN}) proposed in \cite{Li2017}: For this method, the original data sets are transferred using the windowing technique to be in the form of \eqref{eq:transformed_data:window}. The window lengths are set to be $30$, $20$, $30$, and $15$ for the data sets \#1 to \#4, respectively. During test time, we only use one data point corresponding to the last recorded cycle for each engine unit.
    
    We use the same architecture and parameters as in \cite{Li2017}. The network is constructed by stacking four identical convolution layers, each with ten filters of size $10\times1$, followed by another convolution layer with a single filter of size $3\times1$, a Flatten layer, a fully connected layer with $100$ neurons, and a single neuron that outputs the RUL estimation. Zero-padding is used in convolutional layers to keep the data dimension unchanged. Except for the last single neuron that uses a linear activation function, the others use \textit{tanh}. In order to prevent overfitting, the Dropout technique with the rate of $0.5$ is used right after the Flatten layer. The Adam optimizer is used for minimizing the mean squared error (MSE) as a loss function. Different from \cite{Li2017}, for the CQR framework, we replace the MSE loss with the pinball loss function to obtain quantiles, as explained in Section \ref{sec:problem:CQR}. Models are trained for $250$ epochs with a batch size of $512$ samples. The learning rate is set to $0.001$ for the first $200$ epochs and $0.0001$ for the last $50$. 
   
    \item Gradient Boosting (\textbf{GB}): The transformed data set is in the form of \eqref{eq:transformed_data:all}, e.g., the collection of all condition monitoring data from all instances and their corresponding rectified RULs. Once again, only one data point corresponding to the last recorded cycle for each engine unit is used at test time.
    
    The default setting of the \texttt{HistGradientBoostingRegressor} from scikit-learn \cite{Pedregosa2011} is adopted for training the model(s). Like in the case of DCNN, the MSE loss function is replaced by the pinball loss for quantile regression in the CQR framework
    
\end{itemize}
\subsection{Results and discussion}
For each of the C-MAPSS data sets and each learning algorithm, we first divide the training data into (proper) training and calibration sets, such that the data points from the same engine unit end in the same subset. The proportion of the data to be included in the calibration set is fixed at $10\%$. Using the learning algorithm, we train a regression model and eight quantile regression models for the quantile set $\lbrace0.10, 0.15, 0.20, 0.25, 0.75, 0.80, 0.85, 0.90 \rbrace$. This way, we are able to perform conformal prediction using multiple miscoverage rates $\lbrace 0.10, 0.15, 0.20, 0.25\rbrace$. For CP frameworks with normalized non-conformity measure (i.e., SCP+NNM and nex-SCP+NNM), we realize $\sigma$ by training a Random Forest regressor using the default setting of the scikit-learn for \texttt{RandomForestRegressor}. For non-exchangeable methods (i.e., nex-SCP and nex-SCP+NNM), we define the set of weights for every data point $(x_\text{new}, y_\text{new}) $ in the test set as 
\begin{align*}
    w_j = 0.99^{|t(y_\text{new}) - t(y_j)|} \, ,
\end{align*}
$\forall (x_j, y_j)\in \mathcal{D}_\text{calib}$, where $t(y)$ denotes the time index of recording $y$ \cite{FoygelBarber2022}.  

\begin{figure*}[t]
\centering

    \begin{subfigure}[b]{0.75\textwidth}
        \begin{subfigure}{0.4\textwidth}
            \includegraphics[width = \textwidth]{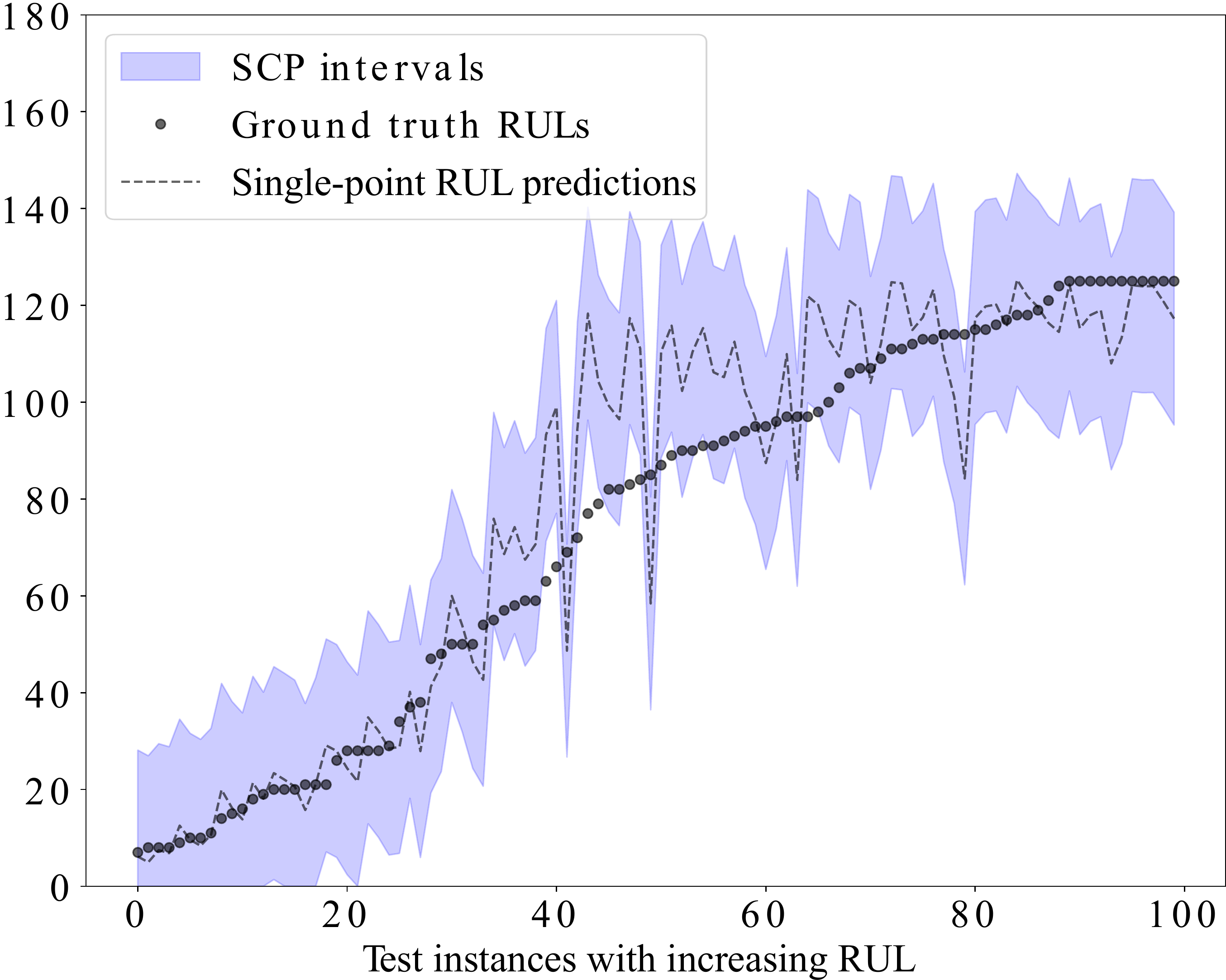}
        \end{subfigure}
        \hfill
        \begin{subfigure}{0.4\textwidth}
            \includegraphics[width = \textwidth]{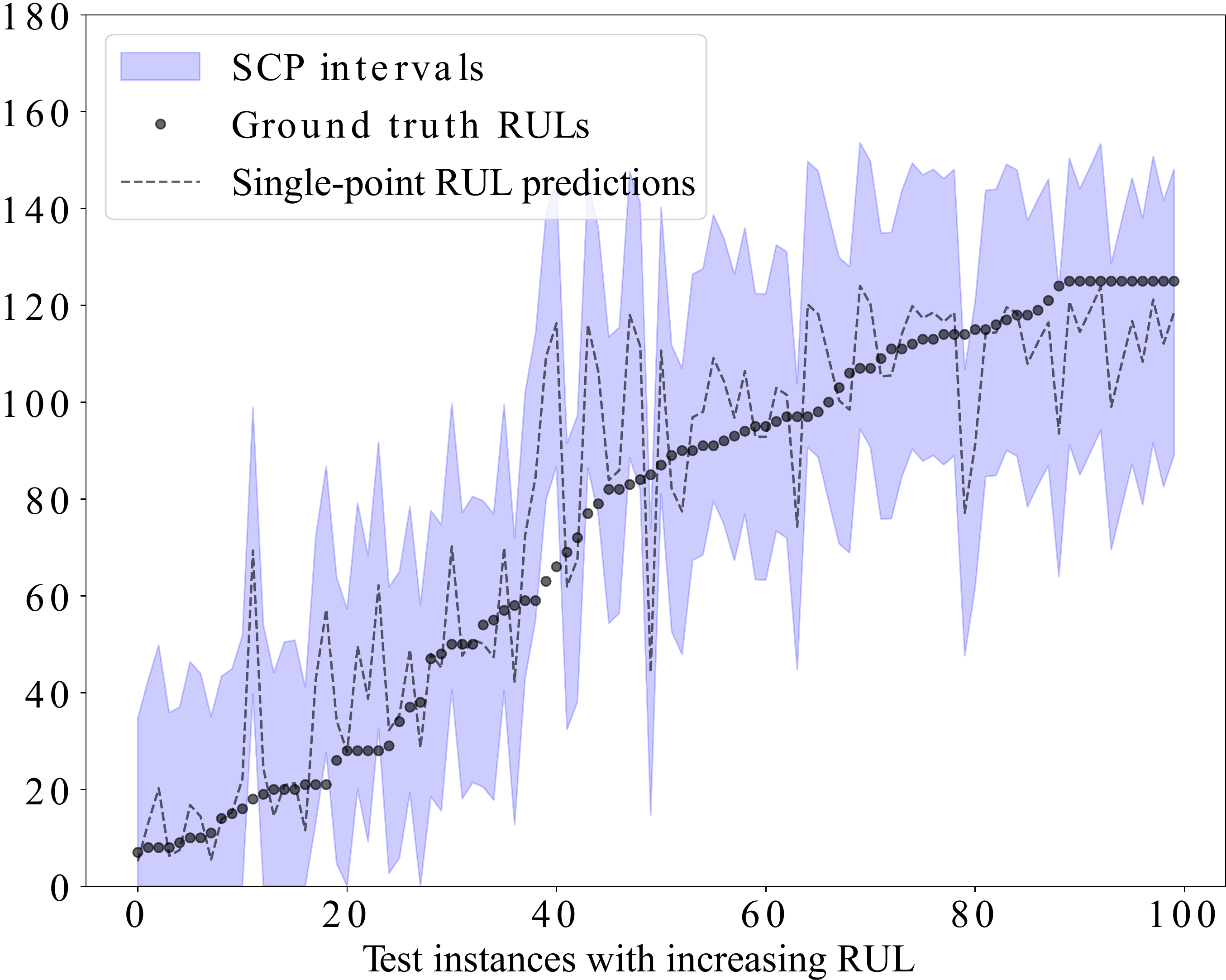}
        \end{subfigure}
    \end{subfigure}
    
    \begin{subfigure}[b]{0.75\textwidth}
        \begin{subfigure}{0.4\textwidth}
            \includegraphics[width = \textwidth]{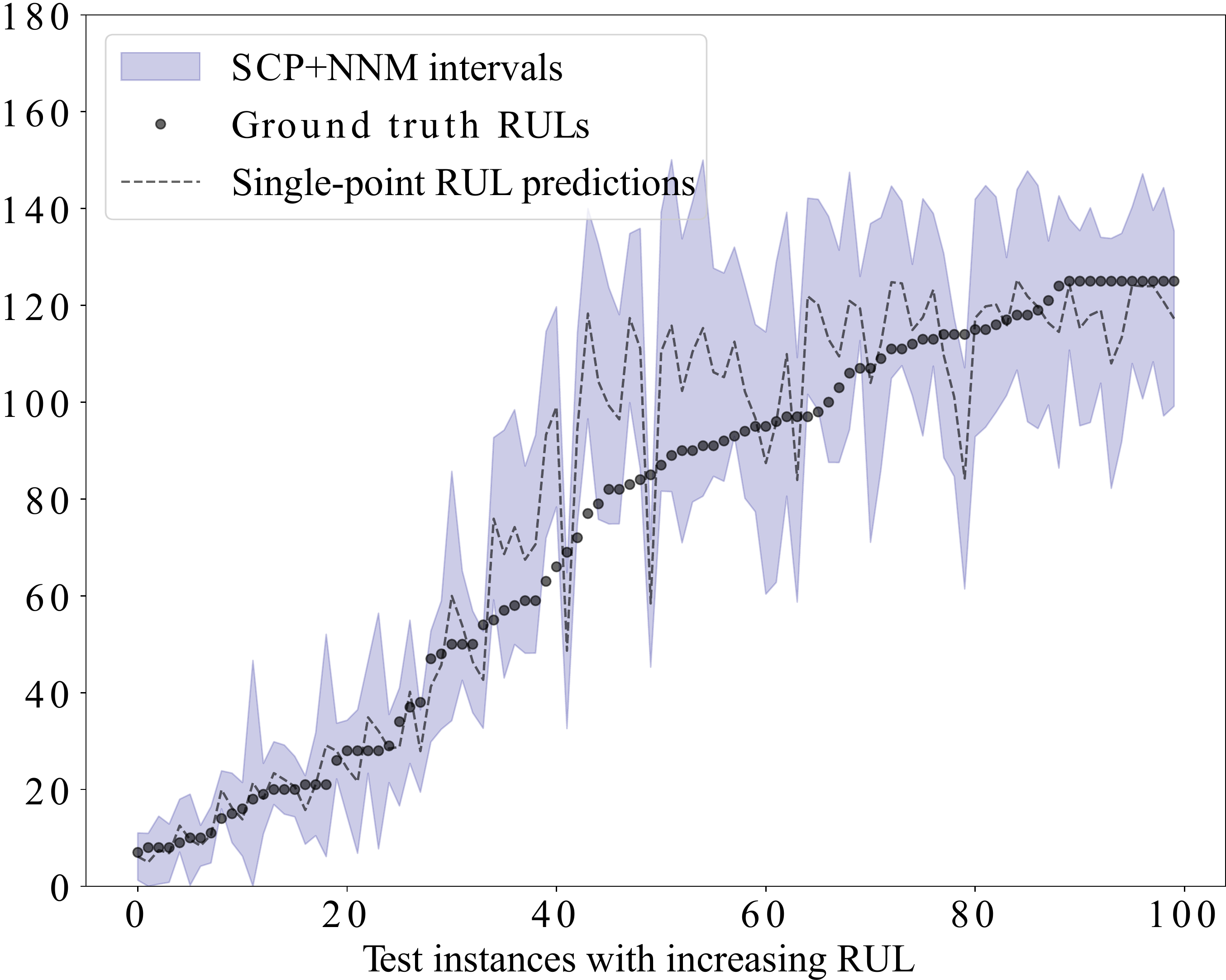}
        \end{subfigure}
        \hfill
        \begin{subfigure}{0.4\textwidth}
            \includegraphics[width = \textwidth]{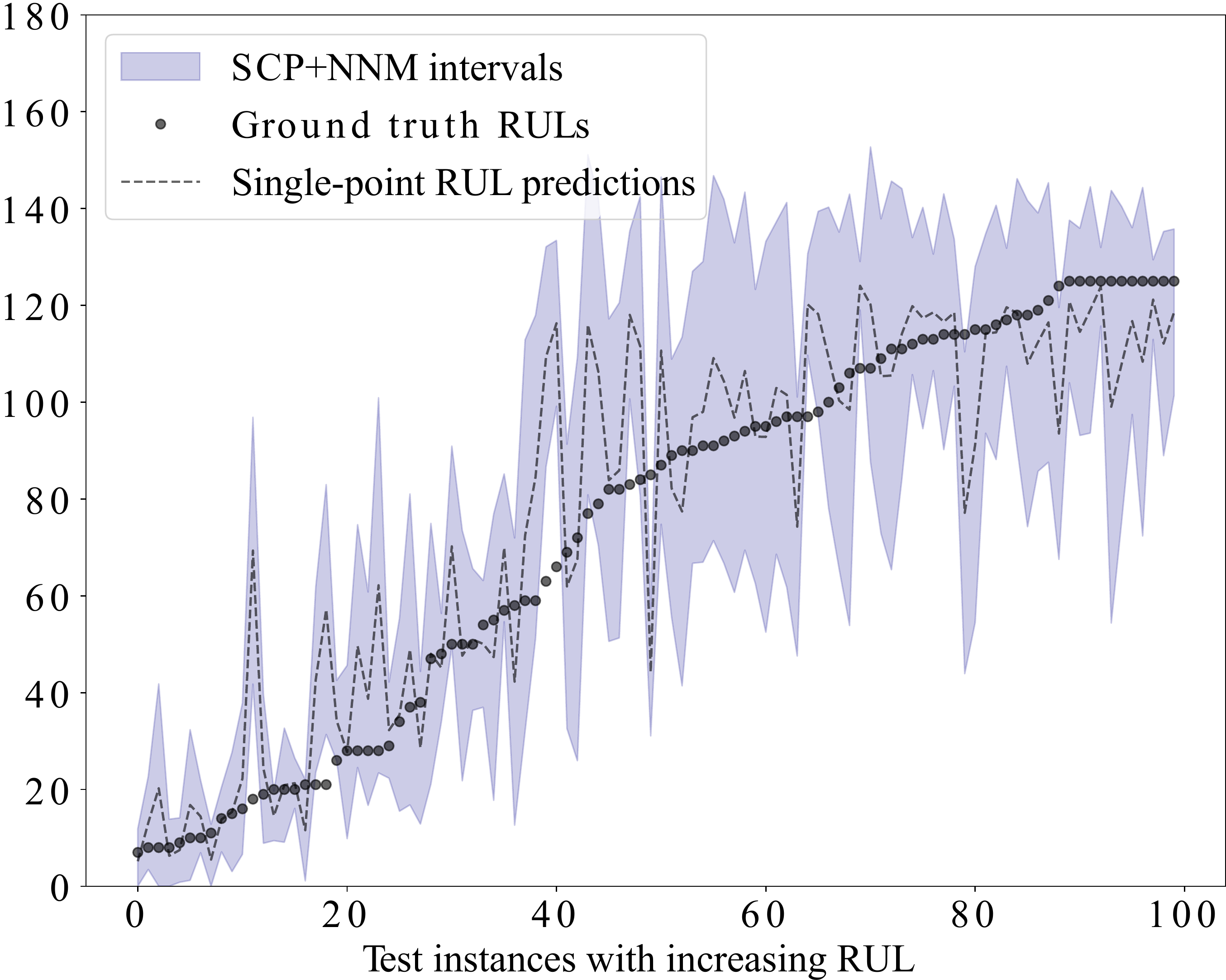}
        \end{subfigure}
    \end{subfigure}
    
    \begin{subfigure}[b]{0.75\textwidth}
        \begin{subfigure}{0.4\textwidth}
            \includegraphics[width = \textwidth]{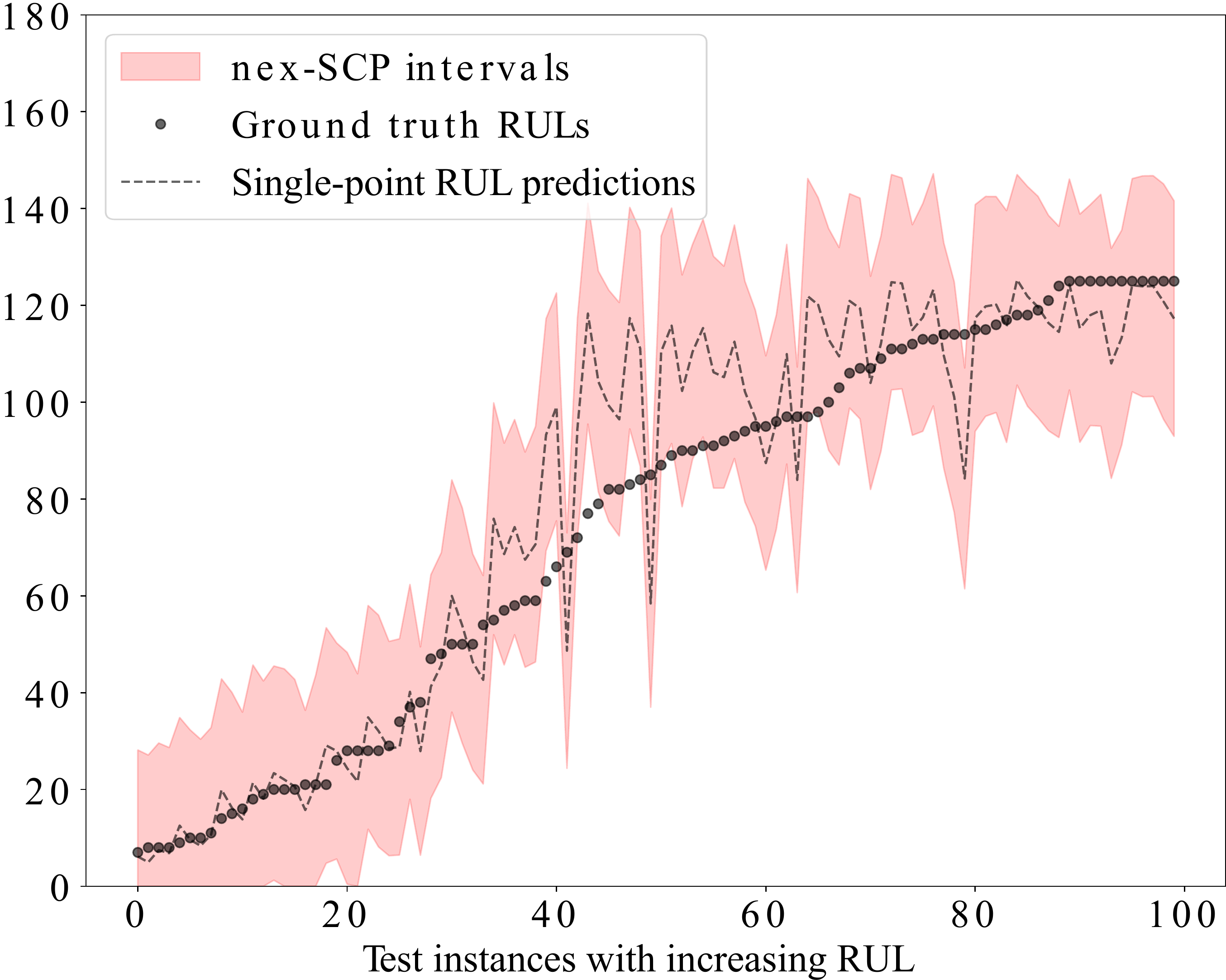}
        \end{subfigure}
        \hfill
        \begin{subfigure}{0.4\textwidth}
            \includegraphics[width = \textwidth]{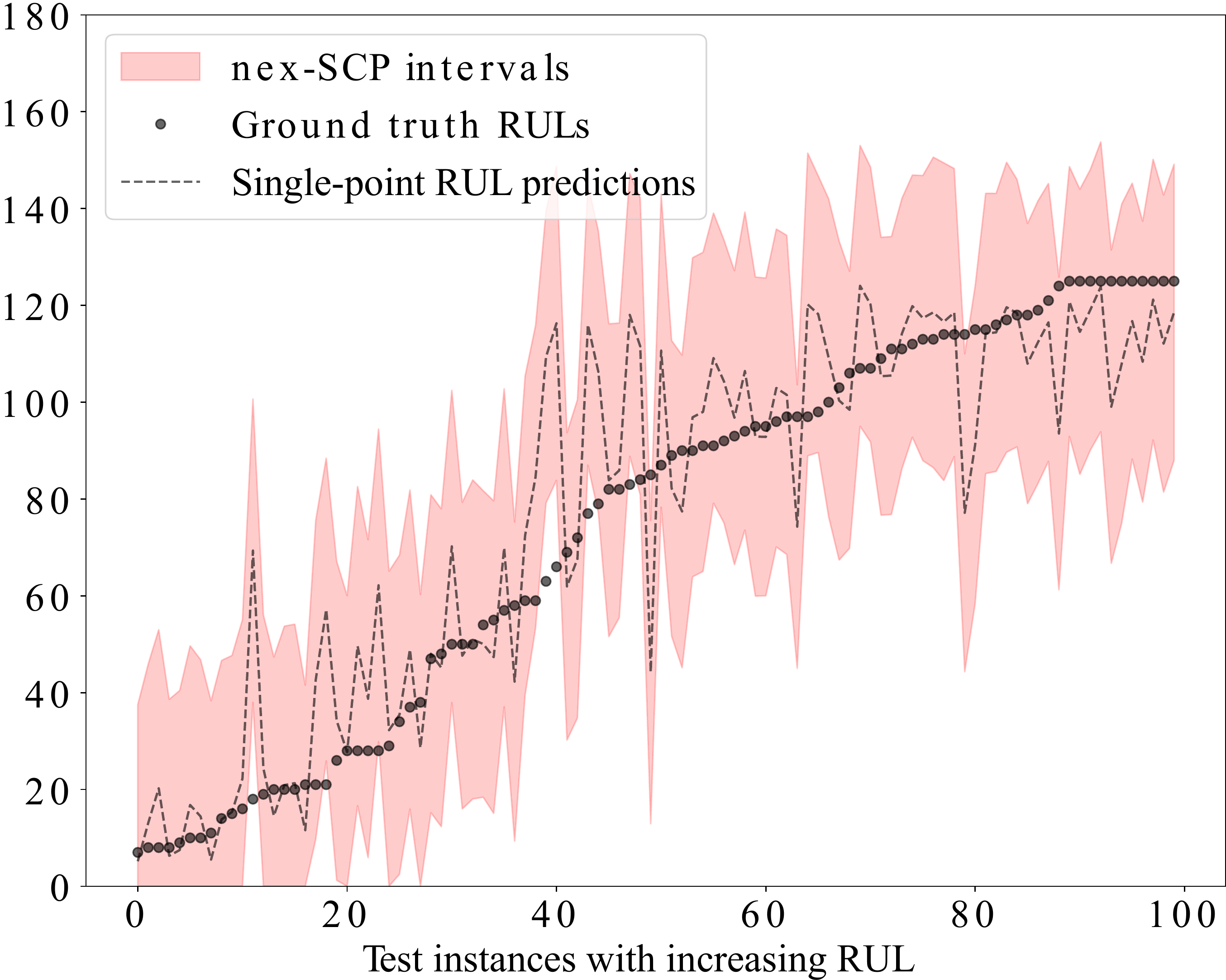}
        \end{subfigure}
    \end{subfigure}
    
    \begin{subfigure}[b]{0.75\textwidth}
        \begin{subfigure}{0.4\textwidth}
            \includegraphics[width = \textwidth]{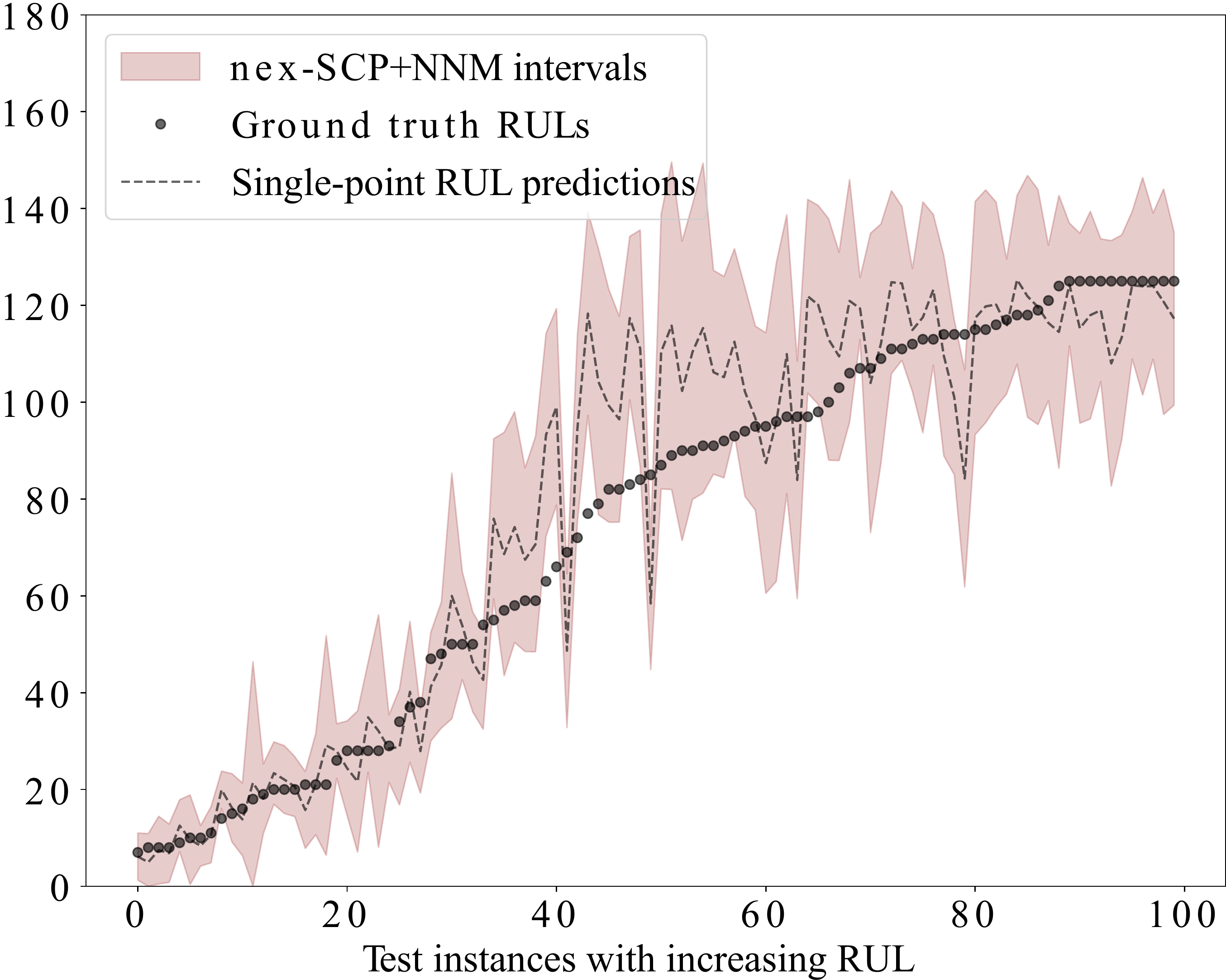}
        \end{subfigure}
        \hfill
        \begin{subfigure}{0.4\textwidth}
            \includegraphics[width = \textwidth]{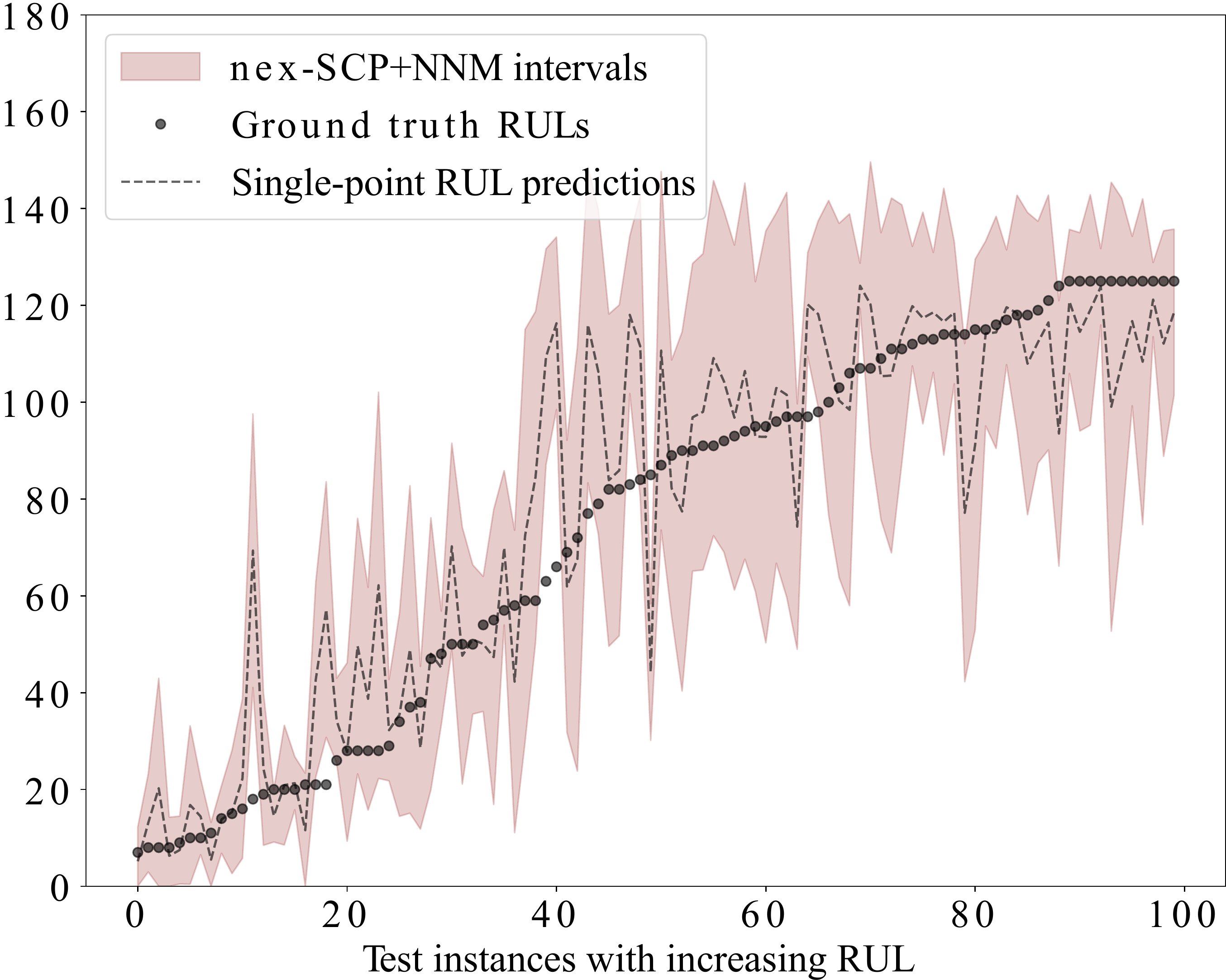}
        \end{subfigure}
    \end{subfigure}
    
    \begin{subfigure}[b]{0.75\textwidth}
        \begin{subfigure}{0.4\textwidth}
            \includegraphics[width = \textwidth]{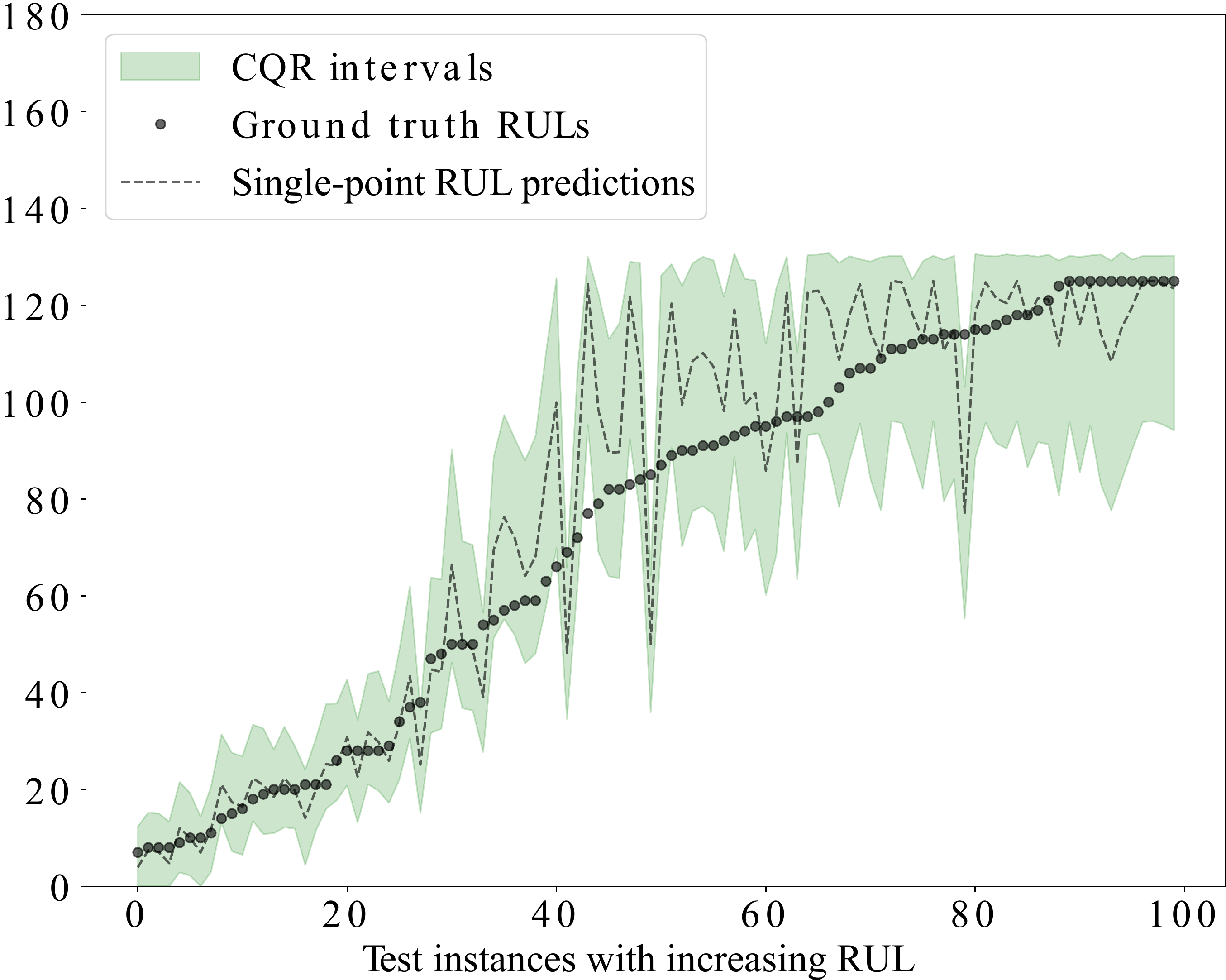}
            \caption{DCNN}
        \end{subfigure}
        \hfill
        \begin{subfigure}{0.4\textwidth}
            \includegraphics[width = \textwidth]{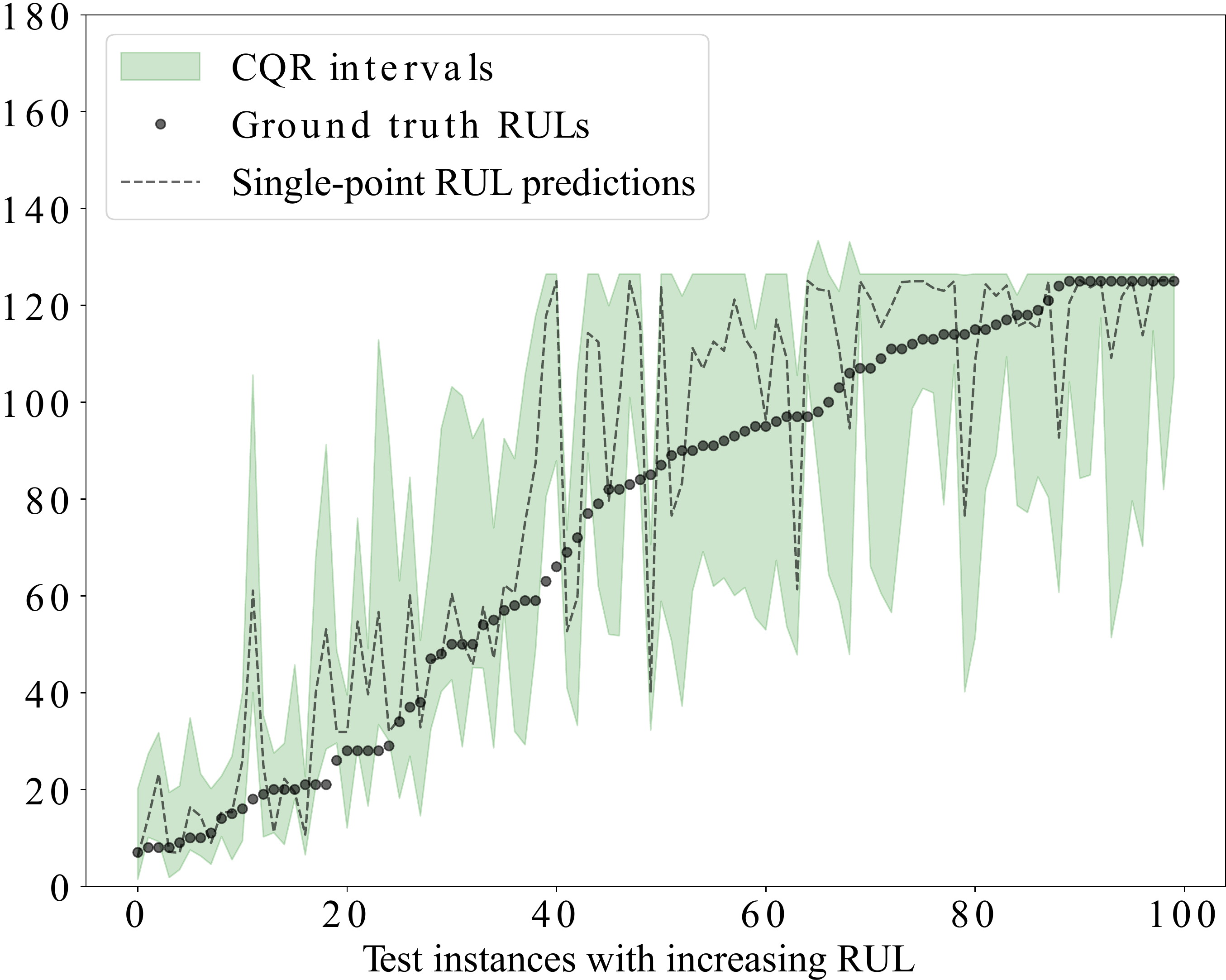}
            \caption{GB}
        \end{subfigure}
    \end{subfigure}
   \caption{Sorted RUL labels of test units of data set \#1 with their predicted intervals from five CP methods.} 
   \label{fig:sorted-RULS:cMAPSS1}
\end{figure*}

Figure \ref{fig:sorted-RULS:cMAPSS1} illustrates the prediction intervals with $\alpha= 0.1$ for the test units of C-MAPSS data set \#1.
By looking at the dashed lines of this figure, one can realize that single-point RUL predictions get more accurate for data points closer to failure times\footnote{The $0.5-$quantile regression model (aka median regression) is used as the single-point RUL estimator in the CQR framework.}. Hence, these data points can be considered easier than those with actual RULs far from zero, and we can expect to observe shorter prediction intervals for them, which is exactly how adaptive methods should behave. Moreover, since DCNN works better than GB in terms of single-point prediction preciseness, its prediction intervals are also shorter on average. 

\begin{figure*}[ht]
\centering
\begin{subfigure}{\textwidth}
    \includegraphics[width = \textwidth]{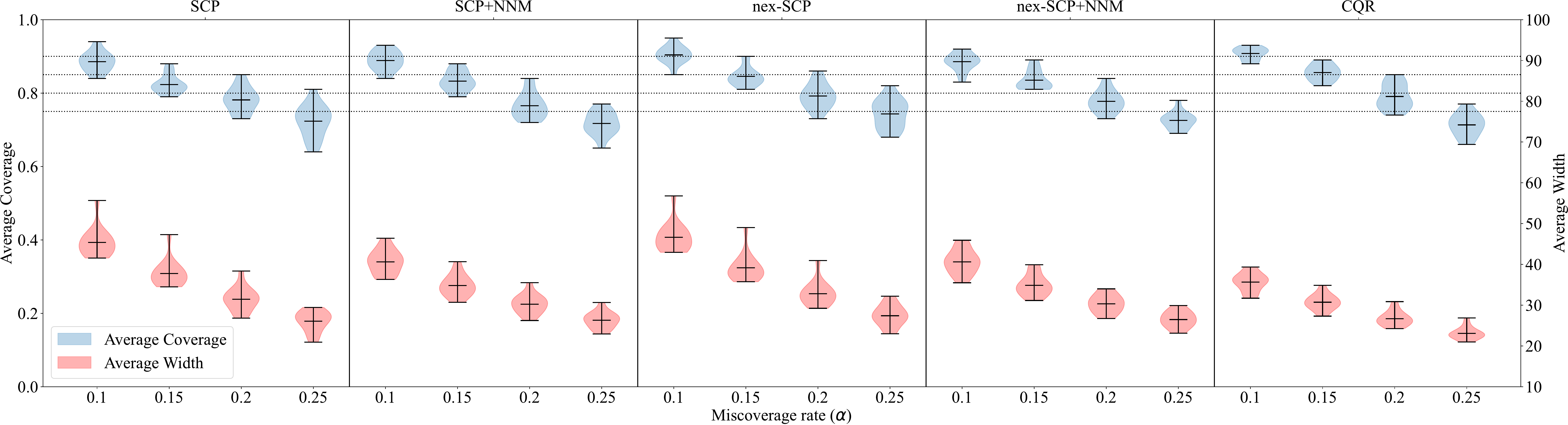}
    \caption{data set $\#1$.}
    \label{fig:cvg-len:DCNN:1}
\end{subfigure}

\begin{subfigure}{\textwidth}
    \includegraphics[width = \textwidth]{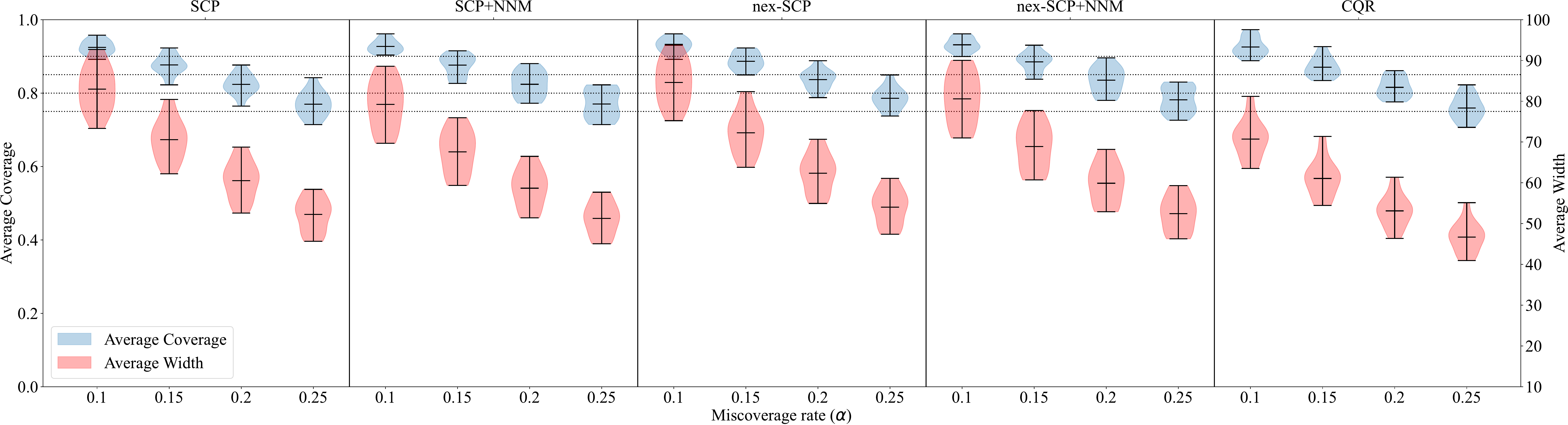}
    \caption{data set $\#2$.}
    \label{fig:cvg-len:DCNN:2}
\end{subfigure}

\begin{subfigure}{\textwidth}
    \includegraphics[width = \textwidth]{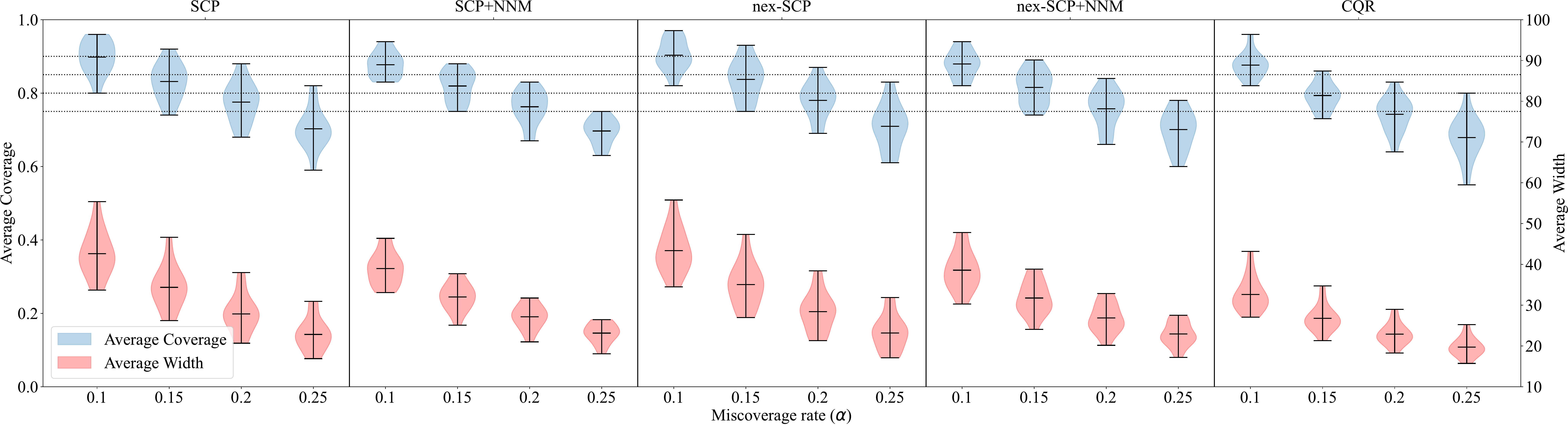}
    \caption{data set $\#3$.}
    \label{fig:cvg-len:DCNN:3}
\end{subfigure}

\begin{subfigure}{\textwidth}
    \includegraphics[width = \textwidth]{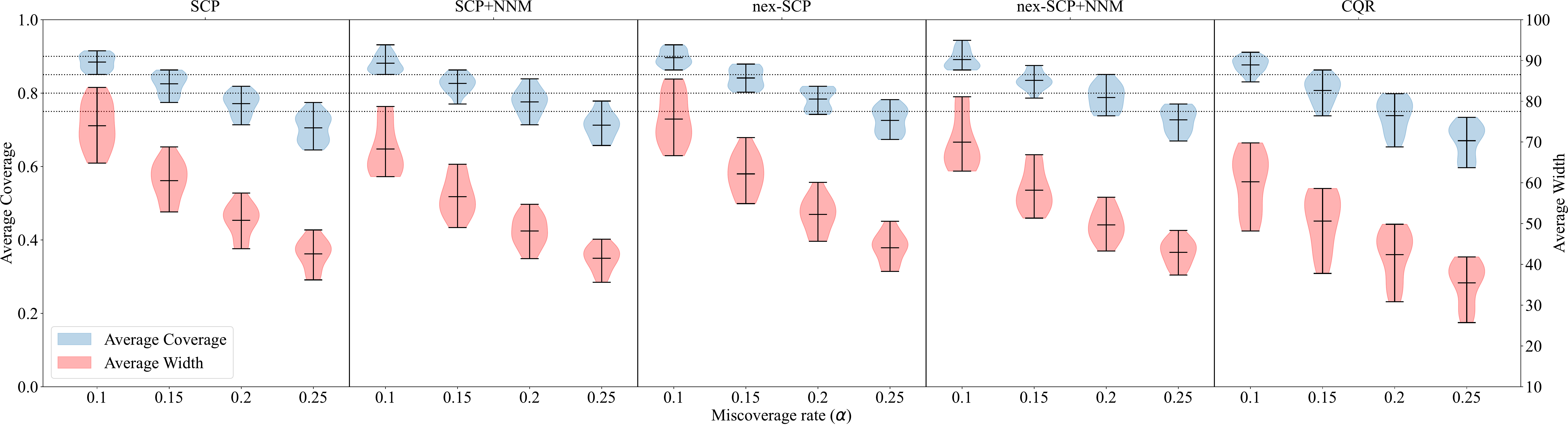}
    \caption{data set $\#4$.}
    \label{fig:cvg-len:DCNN:4}
\end{subfigure}
\caption{Average coverage and the average prediction interval width of different CP frameworks for C-MAPSS data sets, using DCNN as the underlying regression model.}
\label{fig:cvg-len:DCNN}
\end{figure*}

\begin{figure*}[ht]
\centering
\begin{subfigure}{\textwidth}
    \includegraphics[width = \textwidth]{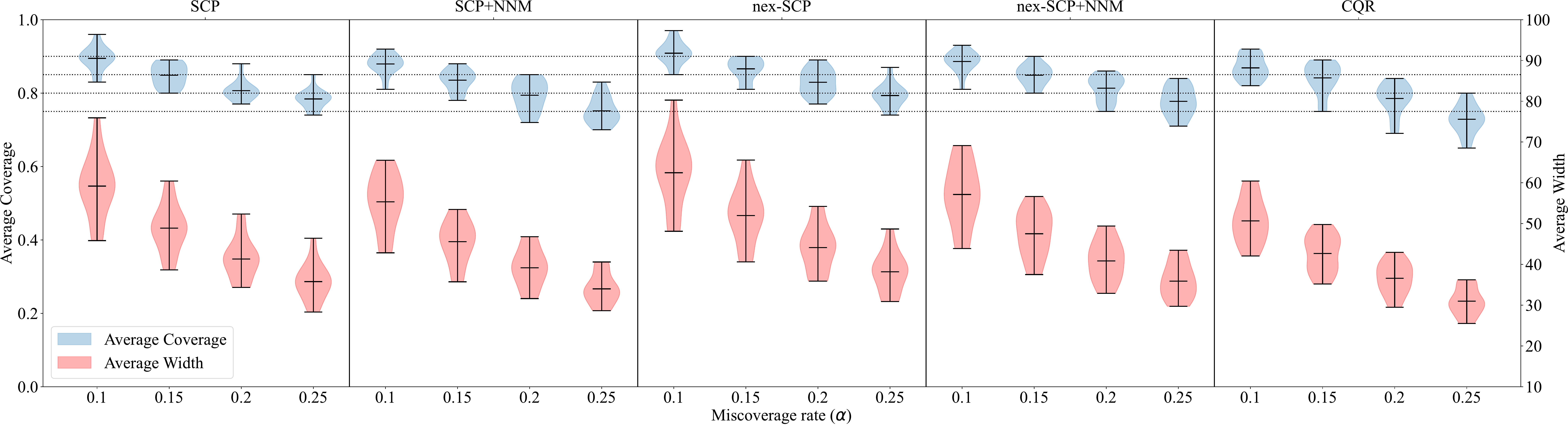}
    \caption{data set $\#1$.}
    \label{fig:cvg-len:GB:1}
\end{subfigure}

\begin{subfigure}{\textwidth}
    \includegraphics[width = \textwidth]{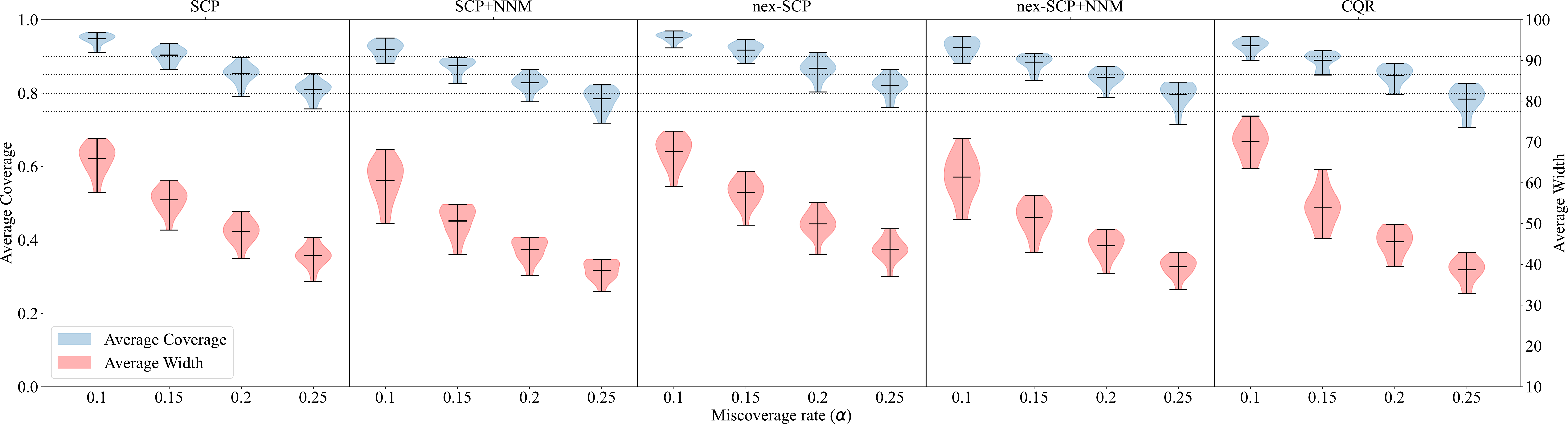}
    \caption{data set $\#2$.}
    \label{fig:cvg-len:GB:2}
\end{subfigure}

\begin{subfigure}{\textwidth}
    \includegraphics[width = \textwidth]{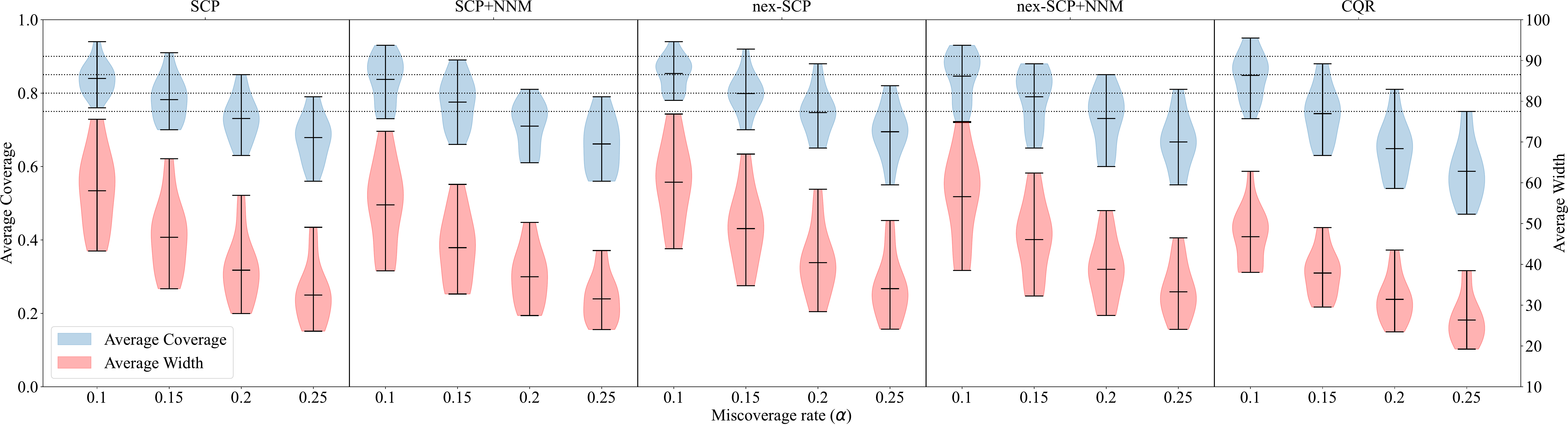}
    \caption{data set $\#3$.}
    \label{fig:cvg-len:GB:3}
\end{subfigure}

\begin{subfigure}{\textwidth}
    \includegraphics[width = \textwidth]{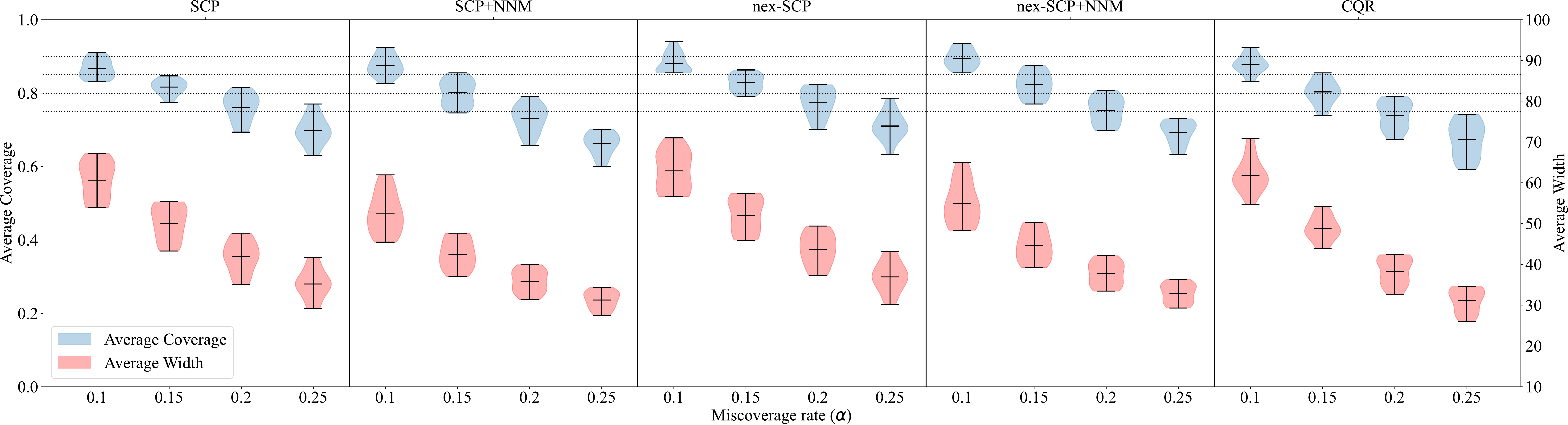}
    \caption{data set $\#4$.}
    \label{fig:cvg-len:GB:4}
\end{subfigure}
\caption{Average coverage and the average prediction interval width of different CP frameworks for C-MAPSS data sets, using GB as the underlying regression model.}
\label{fig:cvg-len:GB}
\end{figure*}

We run the experiments with $15$ random train-calibration splits for each C-MAPSS data set and each learning algorithm. In Fig. \ref{fig:cvg-len:DCNN}, the average coverages and interval widths are shown for all four data sets using DCNN as the underlying learning model. Similar results are provided in Fig. \ref{fig:cvg-len:GB} when GB is the learning model. The horizontal dashed lines indicate the nominal coverage levels (i.e., $0.75$, $0.80$, $0.85$, and $0.90$). On average, non-exchangeable frameworks outperform other frameworks in terms of average coverages, while CQR has the best performance in terms of average interval widths. As expected, the average interval widths decrease as the miscoverage rate increases. Using normalized non-conformity measure often improve the results, especially for nonstationary cases.

\section{Conclusion}
This paper makes the conformal prediction framework amenable to the remaining useful lifetime estimation problem. This allows for specifying the uncertainty of RUL predictions by constructing prediction intervals that include the actual value with a user-defined probability. We reviewed some of the existing CP frameworks and showed how to turn any single-point RUL estimator into an interval predictor. Using deep convolutional neural networks and gradient-boosting algorithms as underlying regression models, we confirmed the validity and evaluated the effectiveness of CP frameworks on the popular C-MAPSS data sets. 

Needless to say, there is still scope for improvement. First, the performance of conformal prediction heavily depends  on the precision of the underlying single-point predictor. Indeed, a more precise model results in lower errors on calibration data and, accordingly, a smaller quantile $q$. This results in shorter prediction intervals while satisfying the coverage property under the exchangeability assumption. Therefore, it is always beneficial to increase the accuracy of regression models for the RUL estimation problem. Moreover, there is also still room for developing better conformal prediction methods for the non-exchangeable case, or even a framework specifically tailored to the RUL estimation problem. 

\section*{Acknowledgment}
This work was supported by the Deutsche Forschungsgemeinschaft (DFG, German Research Foundation) under project number 451737409. 




\end{document}